# A Comparative Study on Parameter Estimation in Software Reliability Modeling using Swarm Intelligence

Najla Akram AL-Saati, Marrwa Abd-AlKareem Alabajee

*Software Engineering Dept., University of Mosul, Iraq.*

*Abstract*-This work focuses on a comparison between the performances of two well-known Swarm algorithms: Cuckoo Search (CS) and Firefly Algorithm (FA), in estimating the parametersof Software Reliability Growth Models. This study is further reinforced usingParticle Swarm Optimization (PSO) and Ant Colony Optimization (ACO). All algorithms are evaluated according to real software failure data, the tests are performed and the obtained results are compared to show the performance of each of the used algorithms. Further more CS and FA are also compared with each other on bases of execution time and iteration number. Experimental results show that CS is more efficient in estimating the parameters of SRGMs, and it has out performed FA in addition to PSO and ACO for the selected Data sets and employed models.

## I. INTRODUCTION

Constructing systems with reliable software has always been atedious task, since the experienced errors in software frequently affect human lives and cost a lot of money year after year. Reliable software can be a very thought-provoking problem. The development of reliable software is mainly hard in cases where there is interdependence among software modules as seen in most of the existing software [1].

Therefore, building reliable software is a major problems, it can be viewed as one of the key elements challenging computer science. Lately, researchers have given this issue a huge attention; many methods were introduced to help system reliability grow.

Software Reliability Growth Models (SRGM) have been proposed for estimating the reliability of software, where sample data (regularly times-to-failure or success data) is employed for estimating parameters of a particular distribution. A *software reliability model* is the mathematical relation found between time consumed by software testing and the accumulative amount of errors discovered [2].

There usually exist two types of models for software reliability namely: Defect Density Models (Predicting software reliability from design parameters), and Software Reliability Growth Models (Predicting software reliability from test data) [1].

A lot of SRGMs have been proposed in the literature; they were used to signify the behavior of detected failures either by times of failures or by the number of failures at particular times [3].

Here, four Swarm Intelligent techniques are to be compared, namely: Firefly Algorithm (FA), Cuckoo Search (CS), Particle Swarm Optimization (PSO), andAnt Colony Optimization (ACO). Which are all to be used in estimating the parameters of the SRGMs; this is carried out using real failure data to show the performance of the employed algorithms. Results will be compared using four models, the Exp (G-O), S-shaped, Power, and the M-O models.

## II. RELATED WORK

SRGMs form a subject of interest for scientists to study and analyze, next are some of these studies. In 2006,Sheta[4]used PSO to estimate the parameters for the exp, power and S-Shaped models.In 2008, Hsu, Huang, and Chen[5], suggested a modified GA with calibrating fitness functions, weighted bit mutation, and rebuilding mechanism for the parameter estimation of SRGMs. In 2009, Yadav and Khan[6], puttaxonomyfor software reliability models reflecting infinite (logarithmic distribution based models) or finite (exponential distribution models) no. of failures.Later in 2010,Satya Prasad, Naga Raju, and Kantam[7], submitted anew model combining imperfect debugging and change-point problems into SRGM.In 2011, Gupta, Choudhary, and Saxena[8], made an analysis using S-shaped model and generalized it by including imperfect debugging and time delay function. Shanmugam and Florence [9] in 2012 compared among best parameter estimation methods and proved ACO to be the best.

Al-Saati and Alabaje[10] in 2013 investigated the use of Cuckoo Search in estimating the parameters for a number of SRGMs. In 2014, Srinivasa Rao [11], proposed models for software prediction to improve failure data, it was taken as a Non-Homogeneous based exponential distribution. Kaur [12] in 2015 employed a tool (CASRE) for measuring reliability. That year also, Wayne and Modarres [13] published a new method to project the reliability growth of a complex continuously OS.

## III. SOFTWARE RELIABILITY GROWTH MODELS (SRGMs)

SRGMs describe the occurrence of failure; they have been established to define software failures by means of a random process and can be used to measure the development status through testing [14].

The estimation of parameters for the Models' equations is carried out using Least Squares Fit or Maximum Likelihood [15]. Each model is capable of providing satisfying results for a precise dataset, but not for all datasets [16].

Failure rates of software system usually decreases with time affected by fault identification and removal. After detecting and repairing faults, SRGMs come to be significant in estimating the improvement of software reliability [17].

Reliability in SRGM will grow with testing time *t*(CPU execution time, man-hours, or days). This is stated in terms of Failure Intensity $\lambda(t)$, or in terms of the Mean Value Function $\mu(t)$ [18].

### A. Classification of SRGMs

SRGMs fall into two types [19]:
- Models described in terms of the failure times of the process. Here, the initial number of faults is unknown but a fixed constant.
- Models described in terms of the number of observed failures. Such as the class of non-homogeneous Poisson process (NHPP) models. The initial number of faults here is a random variable following a Poisson distribution.

In this work, NHPP models are used.

### B. NHPP Model

In NHPP, a proper mean value function is set for the number of failures found until a certain time point. The no. of detected failures up to time (t) can be stated as $(N(t))_{t \geq 0}$[2]. For any finite collection of times $t_1 < t_2 < \ldots < t_n$, the "n" random variables $\{N(t_2)-N(t_1)\} \ldots \{N(t_n)-N(t_{n-1})\}$ are independent. Thus $\{N(t), t>0\}$ has independent increments[20].

If the anticipated number of failures is denoted by $\mu(t)$ in time (t), then $\mu(t)$ is finite, non-decreasing, non-negative and restricted with the boundary conditions. If N(t) has a Poisson probability mass function with parameters $\mu(t)$ as in Eq.(1), then N(t) is called NHPP. Thus, the stochastic behavior of failure can be described by N(t) process [20].

$$P[N(t) = n] = e^{-(\mu(t))} \frac{(\mu(t))^n}{n!} \ldots\ldots\ldots\ldots(1)$$
Where n=0, 1, 2, …, ∞

### C. Models employed

Four models are considered in this work, they are the most commonly and frequently used, they are:

- *Exponential Model (Goel-Okumoto G-O)*
  $\mu(t) = a(1 - e^{-bt})$, $\lambda(t)=abe^{-bt}$

- *The Power Model* (POW)
  $\mu(t) = at^b$, $\lambda(t)=abte^{b-1}$

- *Yamada Delayed S-Shaped Model* (DSS)
  $\mu(t) = a(1 - (1 + bt)e^{-bt})$, $\lambda(t)=ab^2te^{-bt}$

- *Musa-Okumoto Logarithmic Model* (M-O)
  $\mu(t) = a * \ln(1 + bt)$, $\lambda(t)=\frac{ab}{(1+bt)}$

Where
a: is the initial estimate of the total failure recovered at the end of the testing process.
b: is the ratio between the initial failure intensity $\lambda 0$ and total failure.

## IV. SWARM INTELLIGENCE

### A. Cuckoo Search (CS)

In Cuckoo Search, three idealized rules are used to establish a clear description [22]:
- At each time, every cuckoo lays one egg in a randomly chosen nest.

- Only best nests having high quality eggs (solutions) will continue to the next generations;
- The available host nests are fixed in number. A host can discover an alien egg with a probability pa ∈ [0, 1]. When discovered, the host bird can either throw the egg away or dump the nest to build a totally new one in another location.

For simplicity, the third rule can be approximated by a fraction $p_a$ of the *n* nests being replaced by new nests (with new random solutions at new locations). Considering these three rules, the steps of CS can be presented as the pseudo code as follows [21]:

Begin
*Objective function f(x), x = (x1, ..., xd)* <sup>T</sup>
*Generate initial population of* n *host nests* $x_i$ (i = 1, 2, ..., n)
While *(t <MaxGeneration) or (stop criterion)*
*Get a cuckoo randomly by Lévy flights*

*Evaluate its quality/fitness $F_i$*

*Choose a nest among* n *(say,* j*) randomly*
If *($F_i$>$F_j$),*
*Replace* j *by the new solution;*
End
*A fraction ($p_a$) of worse nests is abandoned and new ones are built;*
*Keep the best solutions (or nests with quality solutions);*
*Rank the solutions and find the current best*
*End while*
*Post-process results and visualization*
End

When a new solution $x_i^{(t+1)}$ is generated for the $i^{th}$ cuckoo, a Lévy flight is done as in Eq.(2). Lévy flight is used to conduct a random walk drawn from a Lévy distribution for large steps as in Eq. (3)

$$x_i^{(t+1)} = x_i^{(t)} + \alpha \oplus Lévy(\lambda) \dots\dots\dots\dots (2)$$

*Where*
α: is the step size, usually α = O(1).
⊕: is the entry-wise multiplication

$$Lévy \sim u = t^{-\lambda}, (1 < \lambda \leq 3) \dots\dots\dots (3)$$

This has an infinite variance with an infinite mean. The successive jumps/steps of a cuckoo basically form a random walk which obeys a power-law step-length distribution with a heavy tail.

B. *Firefly Algorithm(FA)*

In order to construct a firefly-inspired algorithm, some characteristics of fireflies have to be idealized as in the following three rules [24]:
- All fireflies are unisex; therefore each firefly is attracted to other fireflies irrespective of their sex.
- Attractiveness and brightness are proportional to each other, so for any two flashing fireflies, the less bright one will move towards the brighter one. Attractiveness and brightness both decrease as their distance increases. If there is no one brighter than other firefly, it will move randomly.
- The firefly's brightness is determined by the view of the objective function.

Hence, the basic steps of the FA can be summarized as the following pseudo code [23]:

Begin
*Objective function $f(x), x = (x_1, \dots\dots, x_d)^T$*
*Generate initial population of fireflies $x_i$ (i =1, 2,..., n)*
*Light intensity $I_i$ at $x_i$ is determined by $f(x_i)$*
*Define light absorption coefficient γ*
While *(t <MaxGeneration)*
For*i = 1 : n all n fireflies*
For*j = 1 : i all n fireflies*
If*($I_j$>$I_i$), Move firefly i towards j in d-dimension*
End If
*Attractiveness varies with distance r via $exp[-\gamma r]$*
*Evaluate new solutions and update light intensity*
EndFor*j*
End For*i*
*Rank the fireflies and find the current best*
End While
*Post-process results and visualization*
End

The movement of a firefly (*i*) attracted to another more attractive firefly (*j*) is computed using the attraction Eq.(4) as shown in Eq. (5)

$$r_{ij} = \sqrt{(x_{i1} - x_{j1})^2 + (x_{i2} - x_{j2})^2} \dots\dots\dots(4)$$

$$x_i = x_i + \beta_0 e^{-\gamma r_{i,j}^2} * (x_j - x_i) + \alpha * \left(rand - \frac{1}{2}\right) \dots(5)$$

*Where*
$x_i$: is the current position of a firefly,
α: is the randomization parameter. α= [0, 1].
*rand*: is a random number generator uniformly distributed in the range of [0, 1].
$\beta_0 = 1$.

## V. DATASETS AND PARAMETER SETTINGS

### A. Datasets

To test the efficiency of CS and FAin parameters estimation of SRGMs, comparisons are made with previous results obtained using PSO (employing3 models) on datasets: Data1, Data2, and Data3[4]. Results are compared again with results obtained using ACO(employing 4 models) on Musa Datasetstaken from the *Data Analysis Center* for Software's Reliability Dataset [25] for Project2, Project3, and Project4.

### B. Parameter Settings

TABLE I shows the settings of the parameters for CS used in this paper and TABLE II shows the parameter settings for FA.

TABLE I
Parameter Settings for Cuckoo Search

| Parameter | Value |
|---|---|
| Lower and Upper bounds for (a) | [0.00001 - 2000] |
| Lower and Upper bounds for (b) | [0.00001 – 1] |
| Number of Cuckoos | 1 |
| Number of Nests | 10 |
| Number of Eggs | 2 |
| Number of iterations (Generations) | 100 |
| Alpha | 0.01 |
| Discovery rate | 0.25 |

TABLE II
Parameter Settings for Firefly Algorithm

| Parameter | Value |
|---|---|
| Lower and Upper bounds for(a) | [0.00001-2000] |
| Lower and Upper bounds for(b) | [0.00001 – 1] |
| Number of fireflies ($n$) | 25 |
| Number of dimensions (d) | 2 |
| Maximum number of generations | 100 |
| Randomization parameter (α) | 0.01 |
| Initial attractiveness ($β_0$) | 1 |
| Absorption coefficient (γ) | 1 |

### C. Evaluation Criterion

In this work,two type of evaluation criteriaare used; the first is the Root Mean Square Error-RMSE given in Eq. (6).The second measure is the Euclidean Distance-ED; its formulation is shown in Eq. (7)

$$RMSE = \sqrt{\frac{1}{N}\sum_{i=1}^{N}(m_i - \mu_i)^2} \quad \ldots\ldots\ldots(6)$$

*Where*

N: is the number of measurements used for estimating model parameters,

$m_i$: is the actual failure number.

$\mu_i$: is the predicted failure number.

$$ED = \sqrt{\sum_{i=1}^{N}(m_i - \mu_i)^2} \quad \ldots\ldots\ldots\ldots(7)$$

*Where*

N, $m_i$, $\mu_i$ is the same as in previous equation (Eq. (6)).

## VI. TESTS AND RESULTS

### A. Comparisons with PSO

The training and testingofFA and CS was done using (70%,30%) training and testing percentages respectively, the same percentages were used by Sheta [4] for Data1, Data2, and Data3, the results are compared for G-O, POW, and DSS Models, TABLES III, IV and V show the comparisons amongFA,CS and PSO for Data1, Data2 and Data3 using RMSE.

Results in TABLE III clarifies that FA was better than PSO and CS only for G-O model, but not for models as CS surpassed other search algorithms.

Results in TABLE IV for Data2 show that CS outperformed others in G-O and POW models. FA was better only for DSS model this time.

As for TABLE V, FA outdid both CS and PSO for POW and DSS models, but not for G-O model.

TABLE III
Comparison using Data1

|  | CS | FA | PSO |
|---|---|---|---|
| G-O | 16.8945 | 15.9041 | 119.4374 |
| POW | 33.6623 | 43.0197 | 152.9372 |
| DSS | 10.9945 | 16.2004 | 26.3015 |

TABLE IV
Comparison using Data2

|  | CS | FA | PSO |
|---|---|---|---|
| EXP(G-O) | 14.2998 | 22.9082 | 80.8963 |
| POW | 56.6807 | 81.5982 | 149.9684 |
| DSS | 11.8833 | 6.9173 | 17.0638 |

### TABLE V
### Comparison using Data3

|         | CS      | FA      | PSO     |
|---------|---------|---------|---------|
| EXP(G-O)| **8.9523** | 10.7637 | 13.6094 |
| POW     | 13.4669 | 12.6660 | 14.0524 |
| DSS     | 15.1916 | 11.8653 | 47.4036 |

### B. Comparisons with ACO

FA and CS were also trained using other datasets and other training percentages, the results were compared with those achieved using ACO which employed the same datasets and (100%) of data for each set for training for the G-O, POW, DSS, and M-OModelsfor Projects 2, 3, and 4. Euclidean Distance was used for performance measuring.TABLES VI, VII and VIII show the results of comparing FA,CS, and ACO for Projects2, 3, and 4 using G-O, POW, DSS and M-O Models.

TABLE VI results for Project2 indicate that FA outperformed ACO but not CS, except for the M-O model where the performance of ACO was the best.

In TABLE VII, FA was better than ACO for all modelsfor Project3. But CS was still better than FA in all of the models except for POW.

TABLE VIII specifies that results were very close between CS and FA, but still FA was better than both ACO and CS for G-O, DSS and M-O models. For POW model, CS outperformed all.

### TABLE VI
### Comparison using Project2

|         | CS         | FA      | ACO     |
|---------|------------|---------|---------|
| EXP(G-O)| **41.7971** | 42.7901 | 60.0371 |
| POW     | **45.9783** | 46.3033 | 52.8854 |
| DSS     | **42.2256** | 42.5206 | 52.8854 |
| M-O     | 41.7732    | 42.2256 | 26.0385 |

### TABLE VII
### Comparison using Project3

|         | CS         | FA      | ACO     |
|---------|------------|---------|---------|
| EXP(G-O)| **21.7256** | 30.4631 | 71.5489 |
| POW     | 15.5885    | 15.3948 | 57.5801 |
| DSS     | **22.4944** | 30.7734 | 57.5801 |
| M-O     | **19.5448** | 20.5183 | 36.1891 |

### TABLE VIII
### Comparison using Project4

|         | CS         | FA      | ACO     |
|---------|------------|---------|---------|
| EXP(G-O)| 25.7682    | 25.7488 | 71.4015 |
| POW     | **28.1951** | 28.5832 | 53.2234 |
| DSS     | 25.7294    | 25.6125 | 53.2234 |
| M-O     | 26.4575    | 26.4196 | 33.1728 |

### C. Further Comparisons between FA and CS

The achieved comparisons between FA,CS, PSO, and ACO show that results of FA is closely to that of CS, and both clearly outperformed PSO and ACO. Here some further comparisons are carrier out between FA and CS, this time the comparisons are based on:

- Speed: the time (T) needed for algorithm to reach to best solution.
- The number of iterations (I) needed by the algorithm to reach the best solution.

TABLES IX, X and XI show the comparisons between the FA and CS for training and testing (70%, 30%) for the first dataset (Data1, Data2 and Data3).

### TABLE IX
### Comparison between FA and CS for Data1

|      | Training 70% |    |       |    | RMSE - Testing 30% |         |
|------|--------------|----|-------|----|---------|---------|
|      | CS           |    | FA    |    |         |         |
|      | T            | I  | T     | I  | CS      | FA      |
| EXP  | 0.047        | 76 | 0.312 | 88 | 16.8945 | 15.9041 |
| POW  | 0.062        | 62 | 0.733 | 97 | 33.6623 | 43.0197 |
| DSS  | 0.063        | 88 | 0.344 | 95 | 10.9945 | 16.2004 |

### TABLE X
### Comparison between FA and CS for Data2

|      | Training 70% |    |       |    | RMSE - Testing 30% |         |
|------|--------------|----|-------|----|---------|---------|
|      | CS           |    | FA    |    |         |         |
|      | T            | I  | T     | I  | CS      | FA      |
| EXP  | 0.047        | 59 | 0.343 | 86 | 14.2998 | 22.9082 |
| POW  | 0.062        | 89 | 0.749 | 92 | 56.6807 | 81.5982 |
| DSS  | 0.046        | 87 | 0.343 | 88 | 11.8833 | 6.9173  |

TABLE XI
Comparison between FA and CS for Data3

|  | Training 70% | | | | RMSE - Testing 30% | |
| --- | --- | --- | --- | --- | --- | --- |
|  | CS | | FA | | | |
|  | T | I | T | I | CS | FA |
| EXP | 0.032 | 45 | 0.172 | 53 | 8.9523 | 10.7637 |
| POW | 0.047 | 29 | 0.328 | 32 | 13.4669 | 12.6660 |
| DSS | 0.047 | 39 | 0.172 | 40 | 15.1916 | 11.8653 |

TABLES XII, XIII, and XIV indicate the comparisons between the FA and CS for training (100%) for the second datasets (Project2, Project3, and Project4) using ED measure.

TABLE XII
Comparing FA and CS for Project2

|  | CS | T | I | FA | T | I |
| --- | --- | --- | --- | --- | --- | --- |
| EXP | 41.7971 | 0.047 | 34 | 42.7901 | 0.328 | 47 |
| POW | 45.9783 | 0.062 | 66 | 46.3033 | 6.131 | 840 |
| DSS | 42.2256 | 0.046 | 29 | 42.5206 | 4.212 | 59 |
| M-O | 41.7732 | 0.047 | 82 | 42.2256 | 0.265 | 85 |

TABLE XIII
Comparing FA and CS for Project3

|  | CS | T | I | FA | T | I |
| --- | --- | --- | --- | --- | --- | --- |
| EXP | 21.7256 | 0.046 | 22 | 30.4631 | 0.249 | 23 |
| POW | 15.5885 | 0.047 | 61 | 15.3948 | 5.07 | 531 |
| DSS | 22.4944 | 0.047 | 57 | 30.7734 | 3.447 | 66 |
| M-O | 19.5448 | 0.032 | 51 | 20.5183 | 0.202 | 64 |

TABLE XIV
Comparing FA and CS for Project4

|  | CS | T | I | FA | T | I |
| --- | --- | --- | --- | --- | --- | --- |
| EXP | 25.7682 | 0.047 | 21 | 25.7488 | 0.327 | 23 |
| POW | 28.1951 | 0.062 | 46 | 28.5832 | 6.521 | 832 |
| DSS | 25.7294 | 0.047 | 65 | 25.6125 | 4.119 | 68 |
| M-O | 26.4575 | 0.047 | 43 | 26.4196 | 0.249 | 47 |

The previous results clearly proved that the CS is much better than FA; this is due to the fact that it requires an obvious less execution times and fewer numbers of Cycles than FA for the two employed data sets with all of the involved models used in this paper.

VII. CONCLUSIONS

In this work, Cuckoo Search and Firefly algorithm were investigated in the estimation of parameters for SRGMs. A number of comparisons were made between CS and FA along with PSO and ACO based on a real failure data. Experimental results showed that CS and FA were very close to each other, both surpassing PSO and ACO.

Further tests were carried out for CS and FA in terms of execution time and number of iterations, results of these tests showed that CS was far better than FA by both execution time and number of iterations.